\documentclass{bmvc2k}
\makeatletter
\input{size11.clo}
\@twosidefalse
\@mparswitchfalse

\renewcommand*{\@evenhead}{\@oddhead}
\makeatother

\title{
Skin Tone in the Wild: A Dataset and Benchmark for 10-tone MST Classification
}

\makeatletter

\renewcommand{\bmv@RenderInst}[1]{{\sffamily\footnotesize #1}}

\renewcommand{\bmv@RenderAuthInstTwoColumn}{%
  \begin{minipage}{\textwidth}
    \raggedright

    \def\bmv@action##1{%
      \bmv@usebox{authnameFN##1}%
      \ifnum##1=1\relax\textsuperscript{,*}\fi
    }%
    \def\bmv@between{\quad}%
    \noindent\makebox[\linewidth][l]{%
      \expandafter\bmv@maplistaux\bmv@auths\bmv@endstop}\par

    \vspace{4pt}

    \def\bmv@action##1{\bmv@usebox{instFN##1}}%
    \def\bmv@between{\quad}%
    {%
      \expandafter\bmv@maplistaux\bmv@insts\bmv@endstop}\par
  \end{minipage}%
}

\makeatother

\addauthor{Vitor Matias}{}{1}
\addauthor{Márcus Lobo}{}{1}
\addauthor{Tiago Novello}{}{2}
\addauthor{João Batista Neto}{}{1}

\addinstitution{ICMC, University of São Paulo}
\addinstitution{IMPA}

\runninghead{Matias et al.}{Skin Tone in the Wild}

\def\eg{\emph{e.g}\bmvaOneDot}

\usepackage{float}
\usepackage{booktabs}
\usepackage{multirow}%

\usepackage[acronym]{glossaries}
\glsdisablehyper
\newacronym{mst}{MST}{Monk Skin Tone}
\newacronym{mste}{MSTE}{Monk Skin Tone example dataset}
\newacronym{fst}{FST}{Fitzpatrick Skin Type}

\newacronym{vit}{ViT}{vision transformer}
\newacronym{dl}{DL}{deep learning}
\newacronym{cv}{CV}{computer vision}
\newacronym{ccv}{CCV}{classic computer vision}
\newacronym{cnn}{CNN}{convolutional neural network}

\newacronym{itw}{ITW}{in-the-wild}
\newacronym{stw}{STW}{skin tone in the wild}

\newacronym{ooacc}{OOAcc}{off-by-one accuracy}
\newacronym{bacc}{bAcc}{balanced accuracy}
\newacronym{wooacc}{wOOAcc}{weighted off-by-one accuracy}

\usepackage{wrapfig}
\usepackage{amsfonts,amsmath,amssymb,amsthm}

\newcommand{\method}[1]{SkinToneNet}
\begin{document}

\maketitle
\makeatletter

\patchcmd{\ps@bmv@c}
  {\textsl{\BMVA@authorOrAnon: \BMVA@title}}
  {}{}{}

\patchcmd{\ps@bmv@c}
  {\textsl{\BMVA@authorOrAnon: \BMVA@title}}
  {}{}{}

\makeatother

\pagestyle{bmv@c}
\thispagestyle{bmv@c}

\begingroup
\renewcommand{\thefootnote}{\fnsymbol{footnote}}
\footnotetext[1]{Vitor Matias:
  \href{mailto:vitorpmatias@usp.br}{vitorpmatias@usp.br}}
\endgroup

\begin{abstract}
Skin tone classification is a fundamental task in computer vision, with applications in quantifying medical features, assessing demographic representativeness, and mitigating algorithmic bias.
However, existing datasets either rely on a restrictive 6-tone Fitzpatrick scale, prohibit training through restrictive licenses, or lack representation of extreme skin tones, while current methods often fail to disentangle true skin tone from in-the-wild imaging conditions, degrading to random-classifier performance.
We introduce \acrfull{stw}, the first fully open-access dataset annotated with the 10-tone \acrfull{mst} scale, containing over 42K images and specifically curated to include underrepresented tones.
Alongside a rigorous evaluation protocol, we propose \method{}, a ViT-based model that leverages global image context for robust in-the-wild generalization.
Experiments show that \method{} achieves $90\%$ weighted off-by-one accuracy, substantially outperforming prior baselines.
We further use \method{} to audit widely used face datasets, including CelebA and VGGFace2, revealing substantial representation imbalances across MST tones.
\end{abstract}

\begin{figure}[h]
    \centering
    \includegraphics[width=\linewidth]{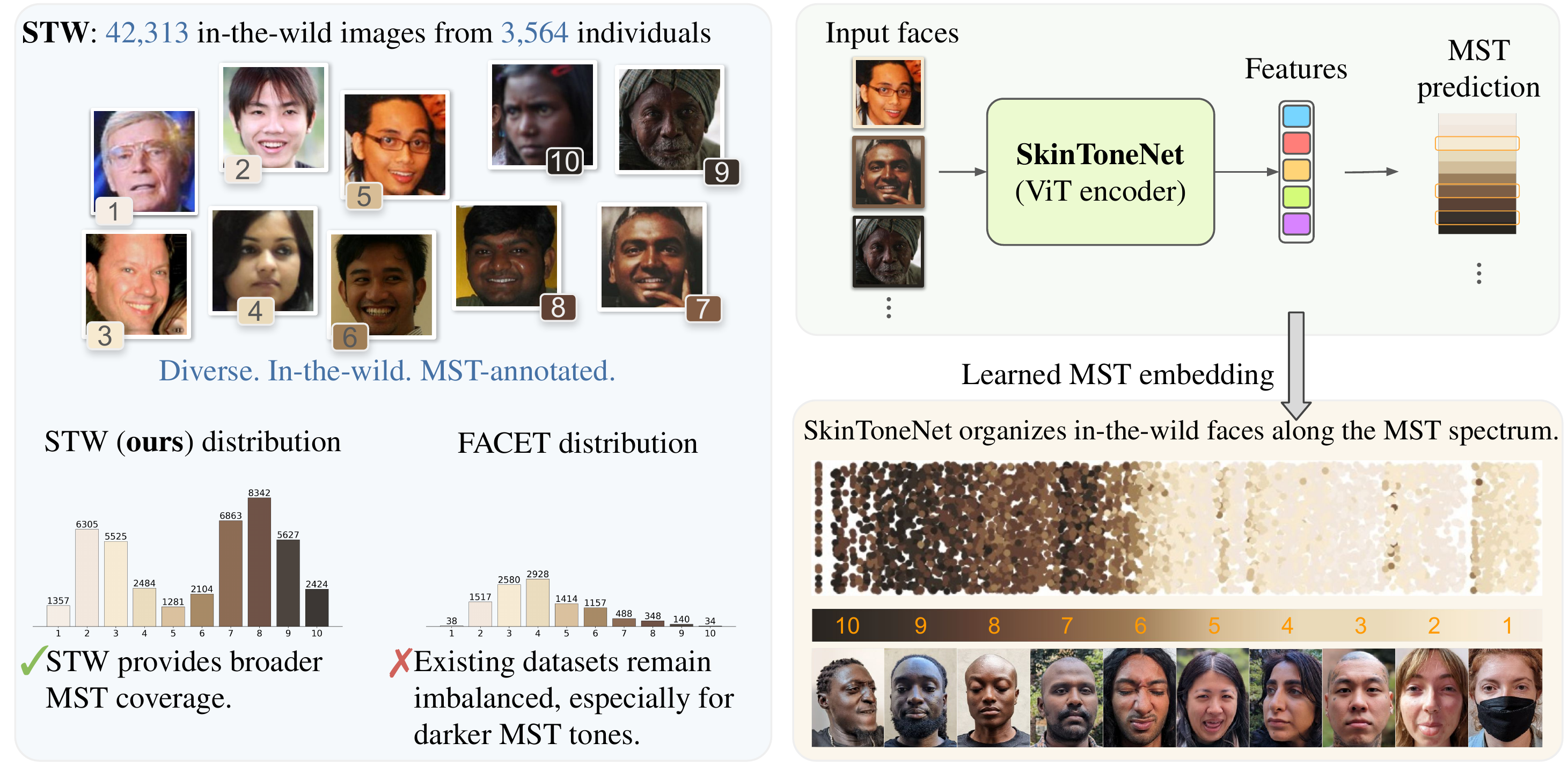}
    \vspace{-0.4cm}
    \caption{
    We introduce \textbf{skin tone in the wild (STW)}, a large-scale dataset annotated on the 10-tone Monk Skin Tone (MST) scale.
    STW provides broader coverage of underrepresented MST tones, especially darker tones, compared to previous datasets such as FACET.
    We also present \textbf{SkinToneNet}, a ViT-based model that takes unconstrained face images as input and predicts MST labels.
    Its learned embedding organizes in-the-wild faces along the MST spectrum, revealing a continuous structure aligned with skin tone.
    }
    \vspace{-0.4cm}
    \label{fig:teaser}
\end{figure}

\vspace{-0.4cm}
\section{Introduction}

Skin tone classification is the task of categorizing human visual skin features and has a wide range of applications in dermatology, cosmetics, content moderation, demographic representativeness analysis, and algorithmic fairness auditing.
Historically, this categorization has often relied on the \acrfull{fst} scale~\cite{fitzpatrick1975soleil}, a 6-tone medical metric originally designed to measure skin reactivity to ultraviolet radiation. However, \acrshort{fst} acts as a medical proxy rather than an accurate visual descriptor of skin tone. To better capture visual skin appearance, recent works have increasingly adopted the \acrfull{mst} scale~\cite{Monk_2019}, a 10-tone scale based on perceived skin color, which provides a more representative~\cite{heldrethMonkRepresentativiness} and granular characterization of human skin tones. 
Despite this progress, automatic skin tone classification remains underexplored in the broader literature on algorithmic bias. Most fairness studies focus on attributes such as gender~\cite{buolamwini2018gender}, age, and ethnicity~\cite{karkkainen2021fairface, Gong_2021_CVPR}, with skin tone analysis frequently reduced to coarse racial proxies (e.g., binary white'' vs. black'' classifications~\cite{tadesse2023skin,borza2018automatic}) or analyzed using the less visually descriptive \acrshort{fst} scale~\cite{groh2021evaluating,hazirbas2021towards}. While skin tone classification raises important sensitivity concerns, it provides a necessary tool for measuring representational disparities and auditing biases in automated vision systems.


Importantly, skin tone classification in unconstrained environments, such as in-the-wild face images and video frames, remains an unsolved challenge due to severe data and algorithmic limitations. First, existing MST-annotated datasets, such as FACET~\cite{FACET} and Casual Conversations v2 (CCv2)~\cite{porgali2023casualconversationsv2dataset}, have restrictive licenses that prohibit model training and therefore limit their use to evaluation.
Moreover, these datasets remain highly demographically imbalanced and provide limited coverage of several MST tones, particularly at the extremes of the scale: 1, 6, 7, 8, 9, and 10.
Other available datasets focus on clinical or dermatological imagery and are labeled with FST, shifting the task toward disease-specific skin-region analysis rather than general in-the-wild skin tone classification~\cite{groh2021evaluating}.
Second, skin tone annotation is inherently subjective and highly affected by imaging conditions~\cite{Barrett_2023}; a subject with an \acrshort{mst}-10 phenotype may visually appear as \acrshort{mst}-3 under intense illumination.
Current methods struggle with such variance, relying either on classical pipelines, such as thresholding heuristics~\cite{groh2021evaluating, kinyanjui2019estimating} and histograms~\cite{kye2022skin} paired with shallow classifiers, while learning-based methods often use outdated convolutional architectures~\cite{rachmadi2015vehicle,robin2020beyond,sobhan2022subject}. Additionally, these approaches are often developed without specialized labeling interfaces or rigorous annotator protocols~\cite{Barrett_2023}, limiting their ability to disentangle skin tone from image acquisition conditions. Furthermore, sophisticated color-specific feature descriptors~\cite{color_desc, picon2011analise} have yet to be thoroughly explored in this domain.

To bridge these gaps, we introduce \textbf{\acrfull{stw}}, a large-scale, fully open-access dataset specifically curated and labeled for 10-tone \acrshort{mst} analysis in unconstrained environments. As shown in \autoref{fig:teaser}, \acrshort{stw} provides broader coverage of the \acrshort{mst} spectrum than existing datasets such as FACET, with improved representation of underrepresented tones, especially darker skin tones. \acrshort{stw} contains 42,313 in-the-wild images from 3,564 individuals and spans the full \acrshort{mst} scale, including tones that are critically underrepresented in prior datasets.
Alongside the dataset, we propose \textbf{\method{}}, a ViT-based model for robust in-the-wild skin tone recognition. \method{} takes unconstrained face images as input and predicts \acrshort{mst} labels while leveraging full-image context to handle challenging lighting, pose, occlusion, and acquisition conditions. Moreover, its learned embedding organizes faces along the \acrshort{mst} spectrum, suggesting that the model captures the ordinal structure of skin tone rather than treating \acrshort{mst} classes as independent categories; see \autoref{fig:teaser}(right).
To ensure rigorous evaluation, we introduce a standardized protocol with strict identity-level (split by individual) splits to avoid data leakage, ordinal-aware metrics such as \acrfull{ooacc}, and both in-domain and out-of-domain evaluation. Through this protocol, we benchmark prior methods and show that they generalize poorly in unconstrained settings, often degrading to near-random performance.
In contrast, {\method{}} achieves robust generalization, reaching approximately $90\%$ \acrshort{ooacc}. Finally, we demonstrate the practical utility of \method{} by auditing widely used face datasets, including IMDB Faces, CelebA, and VGGFace2~\cite{FACET,wang2018devil,liu2018large,cao2018vggface2datasetrecognisingfaces,banerjee2020hallucinating,karkkainen2021fairface,phillips1998feret,huang2008labeled}, revealing systematic imbalances across \acrshort{mst} tones, particularly the underrepresentation of darker skin tones.

In summary, our \textbf{main contributions} are:
\begin{itemize}
    \item \textbf{\acrfull{stw}}, an open-access dataset of 42,313 images from 3,564 individuals labeled on the 10-tone \acrshort{mst} scale and curated via a specialized interface to ensure high inter-annotator agreement.
    \item A \textbf{rigorous evaluation protocol} enforcing strict identity-level splits to eliminate data leakage, together with accuracy (Acc) and off-by-one accuracy (OOAcc) as evaluation metrics.
    \item \textbf{\method{}}, a ViT architecture that leverages full-image context to predict \acrshort{mst} classes, achieving state-of-the-art results and strong out-of-domain generalization.
\end{itemize}

\section{Related Works}
\begin{figure}[h!]
    \centering
    \includegraphics[width=1\linewidth]{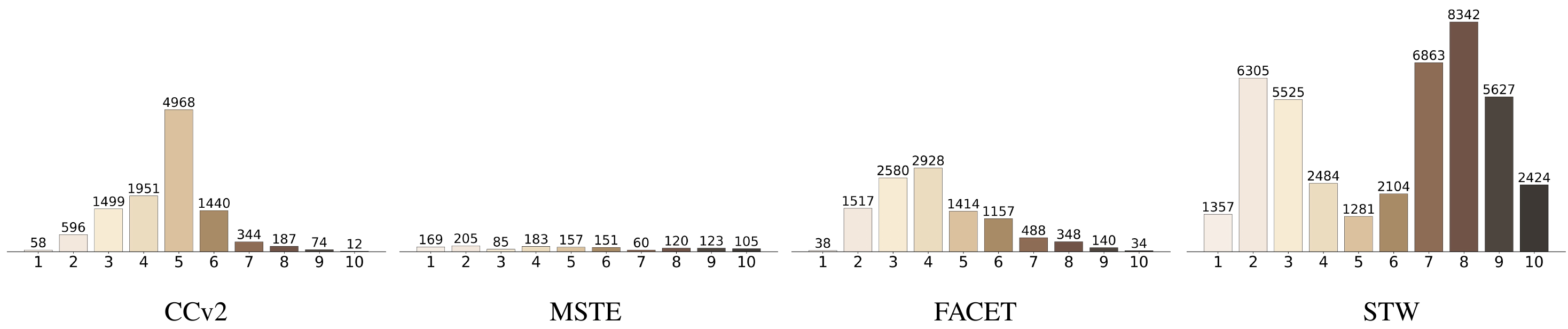}
    \vspace{-0.6cm}
    \caption{\textbf{Dataset distribution along the MST scale.} Compared to existing MST-annotated benchmarks, our skin tone in the wild (STW) dataset provides substantially more in-the-wild samples and vastly superior coverage across the full 10-tone scale, specifically expanding representation for tones at the absolute extremes.}
    \vspace{-0.4cm}
    \label{fig:external_dataset_labels_comparison}
\end{figure}

\paragraph{Skin Tone Datasets and Scales.}
The development of robust skin tone classification models is currently bottlenecked by a scarcity of open-access, large-scale datasets with high-quality annotations. Many existing datasets employ the restrictive 6-tone \acrshort{fst} scale~\cite{hazirbas2021towards, groh2021evaluating, pacheco2020pad, abbi2024scindataset}. Originally designed for UV photosensitivity, FST fails to capture the phenotypic complexity of human skin, particularly for historically marginalized communities, who find it less representative than alternative scales like the \acrshort{mst}~\cite{heldrethMonkRepresentativiness, cook2025colorimetric, groh2022towards, ulrich2025beyond}. While newer datasets such as FACET~\cite{FACET}, Casual Conversations V2 (CCv2)~\cite{porgali2023casualconversationsv2dataset}, and \acrshort{mste} adopt the more inclusive 10-tone MST scale, their restrictive licenses prohibit model training, limiting their use to validation. As shown in \autoref{fig:external_dataset_labels_comparison}, these datasets also remain limited in scale and provide sparse coverage of several tones, particularly at the extremes of the scale. In addition, several prior works rely on proprietary or unreleased datasets~\cite{kye2022skin, kinyanjui2019estimating, boaventura, jagadeesha2023skin, borza2018automatic, robin2020beyond, sobhan2022subject, Pangelinan_2024_WACV}, making direct comparison and reproducibility difficult. In contrast, we introduce \acrfull{stw}, a large-scale, fully open-access dataset for in-the-wild 10-tone \acrshort{mst} classification, designed to support both training and evaluation.

\paragraph{Annotation Procedures.}
Beyond availability, current annotation pipelines lack methodological transparency. Robust dataset creation requires rigorous protocols, multiple human raters, and statistical agreement metrics like Cohen’s $\kappa$ or Krippendorff’s $\alpha$~\cite{Barrett_2023, schumann2023consensus, krippendorff2011computing}. Instead, many datasets rely on automated, pixel-based mappings—such as thresholding Individual Typology Angle (ITA) values via the CIELab color space~\cite{escalaITA}. Mathematically, ITA is defined as $\text{ITA} = \arctan\left(\frac{L^* - 50}{b^*}\right) \times \frac{180}{\pi}$, heavily weighting luminance ($L^*$) and the yellow-blue axis ($b^*$). This formulation completely discards the red-green chromaticity ($a^*$), resulting in systematically erroneous classifications~\cite{Barrett_2023,schumann2023consensus, cook2025colorimetric}. While pixel-based feature extraction performs well in color-calibrated environments utilizing DSLRs or colorimeters~\cite{kye2022skin, robin2020beyond}, such setups are logistically prohibitive for capturing global skin tone diversity due to some skin tones being more easily found in specific parts of the world, for example, in regions like Africa (\acrshort{mst}-9 or \acrshort{mst}-10) or Upper-Scandinavian (MST-1). Consequently, most available facial datasets (skin tone related or not) exhibit severe geographical biases toward European, American, and East Asian populations~\cite{FACET, wang2018devil, liu2018large, karkkainen2021fairface}, drastically increasing the difficulty of curating a balanced, in-the-wild 10-tone dataset.
To address these limitations, \acrshort{stw} is built with a hierarchical annotation protocol and agreement analysis, enabling scalable annotation while improving coverage across the full \acrshort{mst} spectrum

\vspace{-0.4cm}
\paragraph{Skin Tone Classification Models.}
Existing computational approaches for skin tone classification range from classic \acrshort{ccv} pipelines to deep learning (\acrshort{dl}) models. \acrshort{ccv} methods typically segment skin regions (e.g., via YCbCr~\cite{groh2021evaluating} thresholding or RCNNs~\cite{ren2015faster}) and extract color histograms or CIELab averages~\cite{groh2021evaluating, borza2018automatic, kinyanjui2019estimating, ulrich2025beyond}. Recent \acrshort{dl} models utilizing CNNs or ResNets on skin patches report near-perfect accuracies on the FST scale~\cite{sobhan2022subject, robin2020beyond}. However, these high-performance metrics are often artifacts of highly controlled capture environments (e.g., millimeter-sized patches~\cite{kye2022skin}, 18\% gray backgrounds~\cite{borza2018automatic}, colorimeters~\cite{robin2020beyond}) and do not translate to \acrshort{itw} conditions~\cite{kye2022skin, borza2018automatic, Krishnapriya_2022_WACV}. Crucially, both \acrshort{ccv} and \acrshort{dl} methods frequently overlook identity data leakage—where the same subject appears in both training and test splits. Neural networks tend to exploit this leakage via shortcut learning, memorizing, identity-specific artifacts (such as facial geometry or background elements specific to a capture session) rather than the variations that define skin tone. This leads to overfitted non-generalizable models, and explains the $99\%$ accuracy in such a highly subjective problem, with even the authors themselves proclaiming similar reasons~\cite{sobhan2022subject}. While prior work identifies identity leakage, it lacks a solution, out-of-domain evaluation, and explainability~\cite{matias2024enhancing}.
In this work, we establish classical and deep learning baselines under strict identity-disjoint splits, evaluate cross-dataset generalization on CCv2, FACET, and MSTE, and introduce \method{} for robust in-the-wild skin tone recognition.

\section{Skin Tone in The Wild Dataset}\label{sec:dataset}

\begin{figure}[h!]
    \centering
    \includegraphics[width=\linewidth]{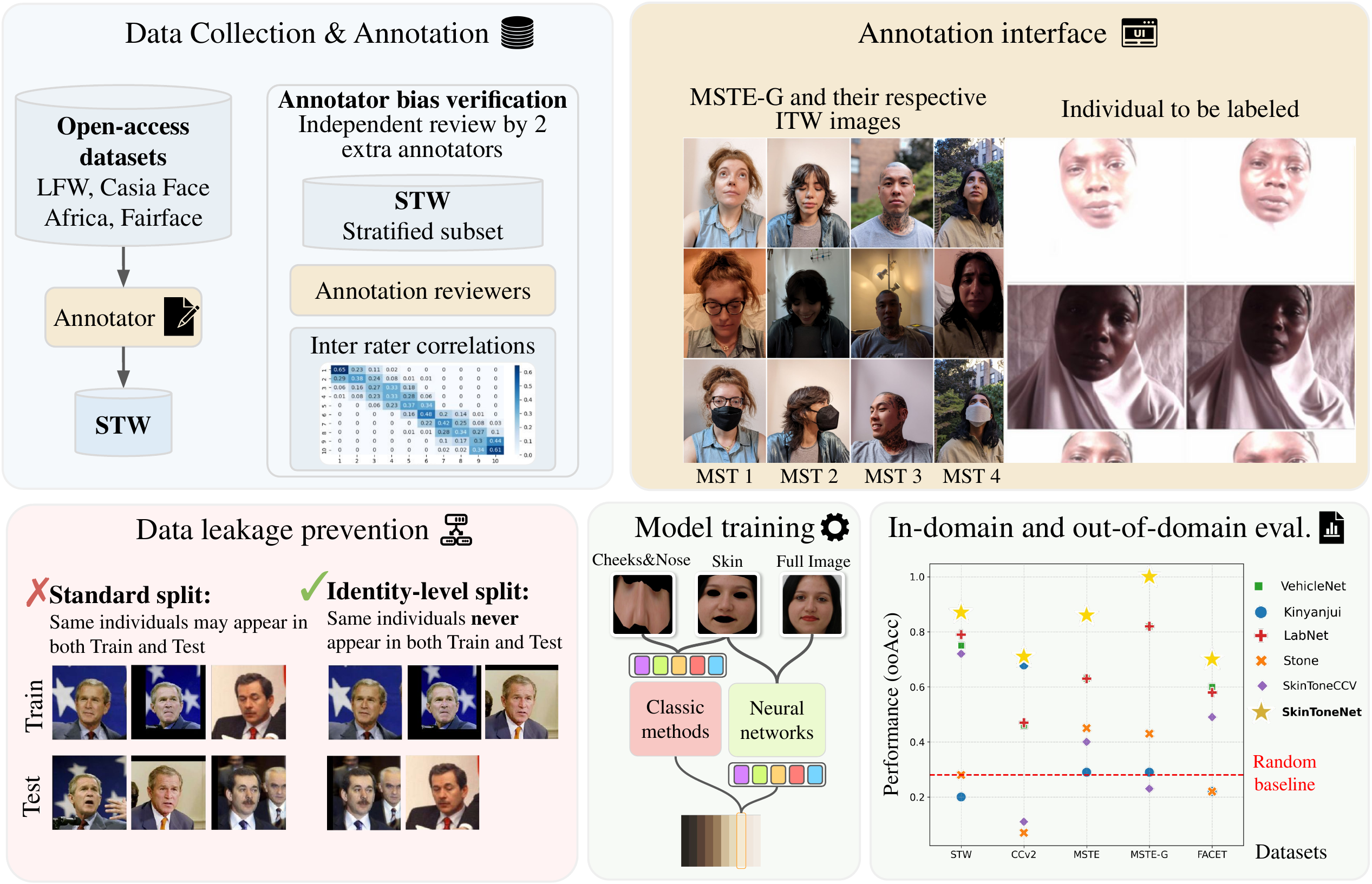}
    \vspace{-0.5cm}
\caption{\textbf{Overview of the STW pipeline.} 
We construct \acrshort{stw} from multiple open-access face datasets and annotate identities on the 10-tone \acrshort{mst} scale (top-left).
Annotators compare each identity against MST reference examples and their in-the-wild image variants (top-right).
We compare standard image-level splits, where the same identity may appear in both training and test sets, with identity-level splits, which prevent identity overlap (bottom-left).
We evaluate both classic and deep learning models using different input representations (bottom-center).
Finally, we benchmark prior and ours models on \acrshort{stw} and external MST-annotated datasets, showing that our method achieves stronger in-domain and out-of-domain generalization (bottom-right).}
\vspace{-0.4cm}
    \label{fig:main_methodology}
\end{figure}

Our goal is to create a large-scale, open-access dataset for in-the-wild skin tone classification on the 10-tone \acrfull{mst} scale~\cite{heldrethMonkRepresentativiness}. While existing MST-annotated datasets provide valuable resources for evaluating skin tone representation, they are often restricted by licenses that prohibit training, limited in scale, or imbalanced across the MST spectrum, particularly at the extremes. To bridge this gap, we introduce \textbf{\acrfull{stw}}, the first fully open-access dataset designed for 10-tone MST classification in unconstrained environments, comprising 42,313 images from 3,564 individuals. \autoref{fig:main_methodology} (top) gives an overview of STW dataset.

\subsection{Data collection}
\acrshort{stw} is built by aggregating images from multiple open-access face datasets and annotating skin tone at the identity level, i.e., all images of the same individual are assigned a single MST label. The source datasets are: 1) Labeled Faces in the Wild (LFW)~\cite{huang2008labeled}, 2) CASIA Face Africa (CFA)~\cite{muhammad2021casia}, 3) CASIA FaceV5~\cite{CFV5}, 4) FEI~\cite{deOliveiraJunior2006FEI}, 5) Faces94 and Faces95~\cite{spacek1995collection}, 6) FairFace~\cite{karkkainen2021fairface}, and 7) CelebA~\cite{liu2018large}.

\begin{figure}[h!]
    \centering
    \includegraphics[width=\linewidth]{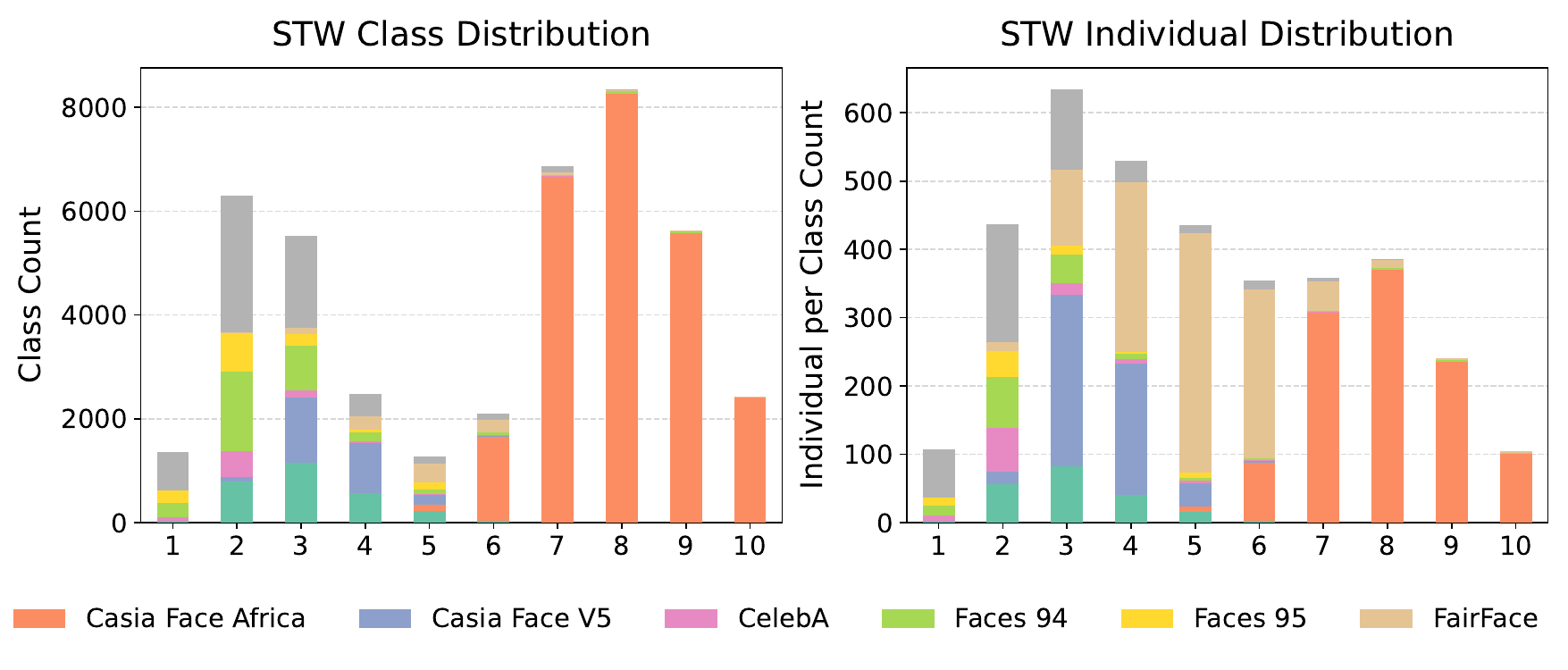}
    \vspace{-0.4cm}
    \caption{\textbf{Distribution of skin tone in the wild (STW).} Standard image-level class distribution (left) and individual-level class distribution (right) across MST classes, with the y-axis denoting the number of images and individuals, respectively. The individual-level distribution shows how targeted aggregation of datasets such as CASIA Face Africa and FairFace improves coverage of underrepresented MST tones.}
     \vspace{-0.3cm}
    \label{fig:dataset_distribution}
\end{figure}

The source datasets were selected to improve coverage across the MST spectrum, with particular emphasis on tones underrepresented in existing benchmarks. Empirical analysis of the aggregated distribution showed that FairFace~\cite{oneto2020fairness} and CASIA Face Africa (CFA)~\cite{muhammad2021casia} were critical for mitigating severe class imbalance, contributing most samples for darker skin tones (MST 7--10), as shown in \autoref{fig:dataset_distribution}. In particular, CFA provided the largest number of individuals in MST classes 7--10, while CASIA Face V5 and FEI mainly contributed samples in MST classes 3 and 4. FairFace further increased coverage of intermediate tones, especially classes 4--6. This targeted aggregation substantially improves individual-level coverage across the MST scale. Although the image-level distribution remains imbalanced, this reflects the practical difficulty of collecting and labeling sufficient in-the-wild samples at the extremes of the scale, especially MST classes 1, 9, and 10.

\subsection{Annotation procedure} 
To label \acrshort{stw}, we designed a split-window interface in which MST reference images, together with low/normal-illumination, and occluded-face examples, are shown on the left, while images of the identity being labeled are shown on the right; see \autoref{fig:main_methodology} (top-right). 
To reduce subjectivity, annotators followed a strict 10-page protocol that prohibited inferences based on race or lighting conditions, focusing solely on visible skin tone. The MSTE-G and their respective ITW images, shown to the left of the Annotation Interface block, were useful when annotators encountered ambiguous images under high or low illumination.

\begin{figure}[h!]
    \centering
    \vspace{0.2cm}
    \includegraphics[width=1\linewidth]{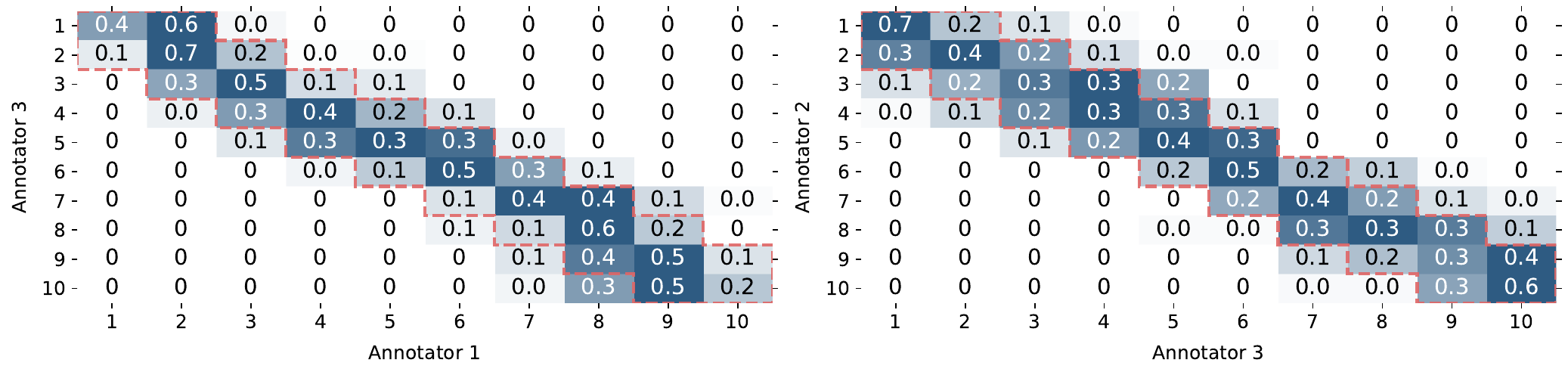}
    \vspace{-0.2cm}
    \caption{\textbf{Inter-annotator confusion matrices.} Correlation between Annotator 1 vs. 3, and Annotator 3 vs. 2. The tight clustering of predictions along the main and adjacent diagonals highlights the ordinal nature of human skin tone perception, resulting in an 88\% off-by-one agreement rate, as shown in red.}
    \label{fig:annotators_comparison}
\end{figure}

The annotation process followed a hierarchical protocol. A single primary expert annotator labeled the full dataset (3,564 individuals) to ensure consistency across samples. To validate label quality, two additional independent annotators labeled a stratified subset of 1,000 individuals (100 per \acrshort{mst} class). Inter-annotator agreement was computed only on this validation subset. As shown in \autoref{fig:annotators_comparison}, the pairwise confusion matrices are highly concentrated along the main and adjacent diagonals. Although the raw average accuracy is 38.8$\%$, the ordinal nature of the task is better captured by an \acrfull{ooacc} of 88$\%$. A reliability analysis further yielded an Intraclass Correlation Coefficient (ICC3) of 0.939 and a Krippendorff’s Alpha of 0.935, indicating excellent inter-annotator agreement~\cite{cicchetti1994guidelines}.

The exact inter-annotator agreement of 38.8\% is consistent with the subjectivity of skin tone perception. Previous dermatological studies report human agreement rates below 40\% for 6-to-8-tone scales~\cite{borza2018automatic, groh2021evaluating,groh2022towards}. In contrast, our \acrshort{stw} reaches 38.8\% exact agreement on a more granular 10-tone scale and 88\% \acrshort{ooacc}, indicating that most disagreements occur between adjacent tones, e.g., MST 4 versus MST 5, rather than as large-scale annotation errors. This supports the suitability of our annotation interface, based on~\cite{matias2024enhancing, Barrett_2023}, for reducing annotation discrepancies.

\section{Experiments}\label{sec:ccv}

We evaluate \acrshort{stw} as a benchmark for in-the-wild skin tone classification and introduce SkinToneNet for robust ITW \acrshort{mst} prediction. Our experiments have four goals: (i) compare prior methods, classical computer vision pipelines, and modern deep architectures under a unified protocol; (ii) quantify the impact of identity leakage by contrasting image-level and identity-level splits; (iii) evaluate in-domain performance and out-of-domain generalization on external \acrshort{mst}-annotated benchmarks; and (iv) demonstrate how SkinToneNet can be used to audit skin tone representation in widely used face datasets.

\subsection{Evaluated methods}

\paragraph{Classical baseline.}
We first define a classical computer vision baseline, SkinToneCCV, based on color descriptors and shallow classifiers. The pipeline, illustrated in \autoref{fig:main_methodology} (bottom-middle), consists of four steps: (1) skin-region segmentation using Mediapipe~\cite{lugaresi2019mediapipe}; (2) color descriptor extraction; (3) histogram re-binarization; and (4) classification with traditional machine learning models.

Unlike prior skin tone classification pipelines~\cite{borza2018automatic,kye2022skin,osman2020multi}, which mainly rely on simple color statistics or thresholding, SkinToneCCV combines hand-crafted color descriptors with shallow classifiers. We evaluate two segmentation strategies: full facial regions, which may include hair, beard, and wrinkles, and localized skin-only regions around the cheeks and nose. Feature extraction uses multi-channel color histograms in RGB, LAB, YCbCr, and HSV~\cite{color_desc,picon2011analise,borza2018automatic,kye2022skin}, together with statistical moments such as variance, skewness, and kurtosis~\cite{osman2020multi}. To capture spatial-color relationships, we further compute border/interior classification and color coherence vectors using a 24-to-6-bit quantization based on the two most significant bits of each channel~\cite{picon2011analise}. The histogram dimensionality is treated as a hyperparameter, reducing 256-bin histograms to compact representations with $128$, $64$, $32$, and $16$ bins. Finally, we benchmark MLP, KNN, SVM, decision trees (DT), and random forest (RF) classifiers, with hyperparameters selected through grid search.

\paragraph{Learning-based baselines.}\label{sec:dl}
To benchmark \acrshort{stw} against modern recognition models, we evaluate both skin-tone-specific architectures and general-purpose visual backbones. We adapt VehicleNet~\cite{rachmadi2015vehicle}, originally proposed for vehicle color classification, to \acrshort{mst} classification, and reimplement LabNet~\cite{robin2020beyond}, originally designed as a ResNet-based skin color regressor, as a classification model. We further evaluate ResNet18~\cite{he2016deep}, DenseNet121~\cite{huang2017densely}, and DINOv2~\cite{oquab2024dinov2learningrobustvisualdinov2}.

\paragraph{SkinToneNet.}
We introduce SkinToneNet, a simple yet strong ViT-based model for in-the-wild \acrshort{mst} classification. SkinToneNet fine-tunes a standard ViT-S/16 backbone on \acrshort{stw} and predicts one of the 10 \acrshort{mst} classes from the full face image, without requiring specialized skin segmentation. Our benchmark shows that, under identity-level splits and diverse in-the-wild supervision, global image context helps handle illumination, pose, occlusion, and acquisition variability. Despite its simplicity, SkinToneNet outperforms classical pipelines, prior skin-tone-specific networks, and generic convolutional baselines, providing a strong reference model for \acrshort{stw} and a practical zero-shot tool for auditing widely used face datasets.

\subsection{Experimental protocol}\label{sec:exp_set}

\paragraph{Metrics.}
Because \acrshort{stw} is naturally imbalanced and \acrshort{mst} labels are ordinal, we report \acrfull{bacc} and weighted off-by-one accuracy (\acrshort{wooacc}) for in-domain evaluation. The weighted off-by-one accuracy is defined as the weighted average of class-specific off-by-one accuracy:
\begin{equation}
  \text{wOOAcc} = \frac{\sum_{i=1}^{C} w_i \,\, \text{Acc}_{\text{off-by-one}}(i)}{\sum_{i=1}^{C} w_i}  
\end{equation}
where $C$ is the number of classes, and $w_i$ is the inverse-frequency weight of class $i$, 
$
    w_i = \frac{N}{C \,\, n_i},
$
The class-specific off-by-one accuracy is
\begin{equation}
    \text{Acc}_{\text{off-by-one}}(i) = \frac{1}{N_i} \sum_{j=1}^{N_i} \mathbb{I}(|y_{i,j} - \hat{y}_{i,j}| \leq 1)
\end{equation}
where $N_i$ is the number of samples in class $i$, $y_{i,j}$ and $\hat{y}_{i,j}$ are the ground-truth and predicted labels for sample $j$, and $\mathbb{I}(\cdot)$ is the indicator function. For external datasets, we report standard accuracy and off-by-one accuracy, since weighted metrics depend on dataset-specific class frequencies.

\paragraph{Train-test splits.} 
We use an 80/20 train-test split and further divide the training set into training and validation subsets for hyperparameter selection. All final hyperparameters are selected using validation performance on \acrshort{stw}; external datasets are reserved exclusively for out-of-domain evaluation. We consider two split protocols, illustrated in \autoref{fig:main_methodology} (bottom-left): a standard image-level split (STD) and an individual-level split (IND). The IND split ensures that all images of the same individual are assigned to a single partition, preventing identity leakage. In contrast, the STD split may place images of the same person in both training and test sets, and is therefore used only as a diagnostic protocol to measure performance inflation. To further quantify leakage, we also evaluate on a custom test set containing 10 held-out individuals per class, whose images are excluded from training.

\paragraph{Input representations.}
To evaluate sensitivity to segmentation, background, and contextual cues, we test three input modes: (i) the full uncropped image; (ii) a segmented facial region, including beard and forehead when present; and (iii) localized skin regions around the cheeks and nose. Classical methods are evaluated on segmented inputs, while learning-based models are evaluated with both segmented and full-image inputs.

\paragraph{Data augmentation.}
To improve robustness to in-the-wild acquisition conditions, we apply spatial transformations, including flips, translations, rotations, and rescaling. We also use brightness and contrast jittering, Gaussian noise, blurring, hue and saturation perturbations, grid shuffling, and coarse dropout. These augmentations discourage reliance on environment-specific artifacts and encourage the model to aggregate evidence across visible skin regions and contextual cues.

\paragraph{Out-of-domain generalization.}
To evaluate cross-dataset generalization, we train all models on the \acrshort{stw} training set and test them on external datasets unseen during training. These include Casual Conversations V2 (CCv2)~\cite{hazirbas2021towards,porgali2023casualconversationsv2dataset}, the \acrfull{mste}~\cite{Monk_2019} and its gold-standard subset \acrshort{mste}-G, and FACET~\cite{FACET}. This protocol tests whether models trained on \acrshort{stw} transfer to benchmarks with different acquisition conditions, annotation protocols, and class distributions.

\paragraph{Previous work evaluation.}
We benchmark prior skin tone classification methods alongside SkinToneCCV and SkinToneNet. When source code is unavailable~\cite{kinyanjui2019estimating,robin2020beyond,rachmadi2015vehicle}, we reimplement the corresponding architectures or feature extraction pipelines. To isolate the contribution of each method from preprocessing differences, prior methods are evaluated using their best compatible segmentation setting; in our experiments, this corresponds to segmented skin inputs.

\paragraph{Implementation details.}
All learning-based models, including SkinToneNet, are implemented in PyTorch and trained on a RTX A5000 GPU. Hyperparameters are selected through grid search over batch size, augmentation strategy, split protocol, optimizer, learning-rate schedule, and loss functions. We evaluate standard cross-entropy (CE), class-weighted CE, and an ordinal CE variant defined as
$
(1+(\hat y-y)^2/C)\cdot \text{CE},
$
with $C=10$ classes. For pretrained architectures, we consider linear probing, partial fine-tuning, and full fine-tuning. Input images are resized to the required resolution of each architecture and normalized using \acrshort{stw} channel statistics. For SkinToneCCV, feature extraction and classical classifiers are implemented with OpenCV and scikit-learn. The methods of \citet{groh2021evaluating} and \citet{kinyanjui2019estimating} are reimplemented in PyTorch for efficient inference.

\subsection{Benchmark results}\label{sec:results}

\begin{figure}[h]
    \centering
    \includegraphics[width=\linewidth]{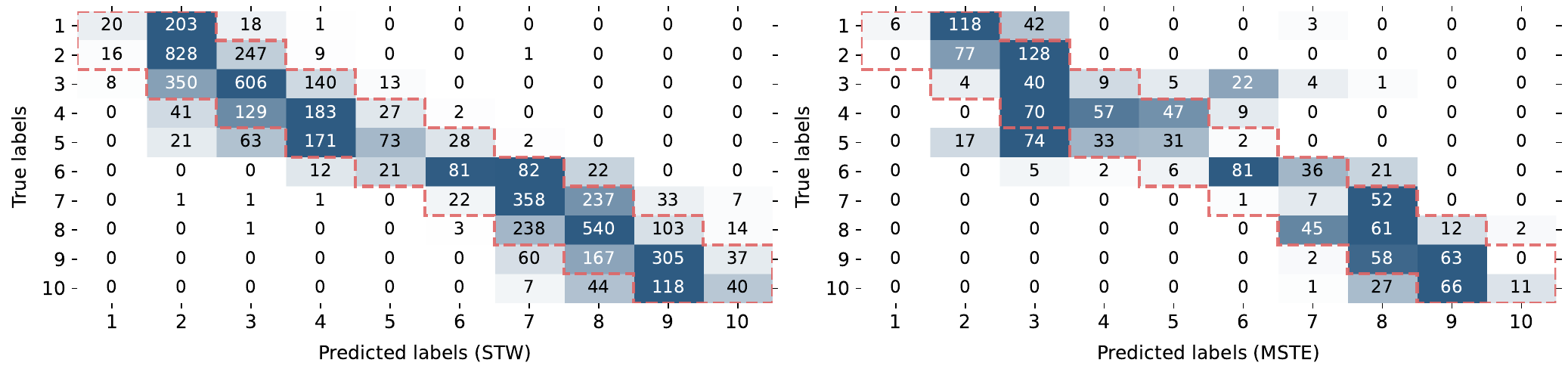}
    \vspace{-0.5cm}
    \caption{\textbf{SkinToneNet confusion matrices.} Evaluation on the in-domain STW test set (left) and the out-of-domain MSTE dataset (right). The strong diagonals and tight clustering of misclassifications along adjacent classes highlight the model's robust ordinal predictions and high off-by-one accuracy (\acrshort{ooacc}), as shown in red.}
    \label{fig:conf_matrix_example}
\end{figure}

\paragraph{SkinToneNet and baselines.}
\autoref{tab:compact_results} summarizes in-domain performance on \acrshort{stw} and cross-dataset generalization to external \acrshort{mst}-annotated benchmarks. The results show that \acrshort{stw} is challenging: most prior 10-class methods perform only modestly above random, while SkinToneCCV fails to generalize reliably under in-the-wild variation. Among 10-class methods, \textbf{SkinToneNet achieves the best results on every dataset}. On \acrshort{stw}, SkinToneNet improves over the strongest prior 10-class baseline by +5 points in accuracy and +8 points in \acrshort{ooacc}. Furthermore, on the out-of-domain generalization task, it improves over LabNet by +15/+23 points on MSTE and +10/+24 points on CCv2, and reaches 100\% \acrshort{ooacc} on MSTE-G. These gains are also reflected in the confusion matrices in \autoref{fig:conf_matrix_example}, where errors concentrate along adjacent \acrshort{mst} classes, consistent with the ordinal structure of the task. Compared with SkinToneCCV, SkinToneNet is substantially more robust across external datasets, with gains of up to 30 points in accuracy and 60 points in \acrshort{ooacc}. Additionally, we also report methods with a smaller skin tone scale span (2, 6, and 8) for completeness, although they are not directly comparable to the 10-tone \acrshort{mst} task, since coarser label spaces can inflate \acrshort{ooacc}, they underperform when compared to SkinToneNet. Overall, \acrshort{stw} defines a rigorous in-the-wild benchmark, with SkinToneNet as its strongest 10-class reference model.
\begin{table}[h!]
\centering

\resizebox{\textwidth}{!}{%
\begin{tabular}{c|c|ccccc|c}
\hline
Models\textbackslash Dataset & \begin{tabular}[c]{@{}c@{}}N  of \\ Cls.\end{tabular} & \multicolumn{1}{l}{\textbf{STW {\footnotesize(Ours)}}} & \multicolumn{1}{l}{CCv2} & \multicolumn{1}{l}{MSTE} & \multicolumn{1}{l}{MSTE-G} & FACET & \multicolumn{1}{l}{Random} \\ \hline
\citet{tadesse2023skin} & 2 & 0.89 / 1.00 & 0.53 / 1.00 & 0.60 / 1.00 & 0.56 / 1.00 & 0.56 / 1.00 & 0.5 / 1.00 \\
\citet{groh2021evaluating} & 6 & 0.19 / 0.46 & 0.25 / 0.74 & 0.23 / 0.56 & 0.41 / 0.93 & 0.19 / 0.23 & 0.17 / 0.44 \\
\citet{kinyanjui2019estimating} & 8 & 0.12 / 0.20 & 0.17 / 0.68 & 0.15 / 0.29 & 0.15 / 0.29 & 0.08 / 0.22 & 0.13 / 0.34 \\ \hline
stone~\cite{pinastonecascorenealejandro} & 10 & 0.12 / 0.28 & 0.14 / 0.07 & 0.14 / 0.45 & 0.20 / 0.43 & 0.06 / 0.22 & 0.10 / 0.28 \\
VehicleNet~\cite{rachmadi2015vehicle} & 10 & 0.38 / 0.75 & 0.16 / 0.46 & 0.19 / 0.63 & 0.24 / 0.82 & 0.22 / 0.60 & 0.10 / 0.28 \\
LabNet~\cite{robin2020beyond} & 10 & 0.36 / 0.79 & 0.16 / 0.47 & 0.25 / 0.63 & 0.47 / 0.82 & 0.21 / 0.58 & 0.10 / 0.28 \\
SkinToneCCV {\footnotesize(Ours)} & 10 & 0.34 / 0.72 & 0.13 / 0.11 & 0.14 / 0.40 & 0.07 / 0.23 & 0.18 / 0.49 & 0.10 / 0.28 \\
\textbf{SkinToneNet {\footnotesize(Ours)}} & 10 & \textbf{0.43 / 0.87} & \textbf{0.26 / 0.71} & \textbf{0.40 / 0.86} & \textbf{0.64 / 1.00} & \textbf{0.27 / 0.70} & 0.10 / 0.28 \\ \hline
\end{tabular}}
\caption{\textbf{In-domain and out-of-domain benchmark comparison.} Each entry reports Acc/\acrshort{ooacc}. SkinToneNet outperforms prior methods, classical baselines, and learning-based methods across STW and external MST-annotated datasets. Cross-dataset evaluation uses Acc and \acrshort{ooacc}, since balanced and weighted metrics depend on dataset-specific class distributions.}
\label{tab:compact_results}
\end{table}

\paragraph{Impact of splits and input.}
The results above use the individual-level split. To quantify the effect of identity leakage, we compare the Standard (STD) and Individual-level (IND) splits in \autoref{tab:image_ind}. Under the STD split, both classical and learning-based models achieve artificially high performance, reaching near-saturated \acrshort{wooacc}. However, evaluation on the held-out Custom Test set reveals a severe drop: SkinToneCCV approaches random-level behavior, while SkinToneNet loses roughly 40 points in \acrshort{wooacc}.
\begin{table}[h!]
\centering
\small
\begin{tabular}{lcccccc}
\hline
\multirow{2}{*}{Model} & \multirow{2}{*}{Split} & \multirow{2}{*}{Seg}   & \multicolumn{2}{c}{STW - Test} & \multicolumn{2}{c}{STW - Custom Test}             \\
&&& \acrshort{bacc}& \acrshort{wooacc}& \acrshort{bacc}& \acrshort{wooacc}\\\midrule
SkinToneCCV& \multirow{4}{*}{Standard}   & Skin & 0.889 & 0.941 & 0.168 & 0.442 \\
SkinToneCCV& & Cheeks-Nose & 0.879 & 0.929 & 0.196 & 0.434 \\
SkinToneNet & & Skin & 0.836      & 0.924       & 0.209 & 0.657 \\
SkinToneNet & & Full image & 0.905      & 0.939       & 0.356 & 0.541 \\\midrule
SkinToneCCV & \multirow{4}{*}{Individual} & Skin & 0.331 & 0.682 & 0.322 & 0.656 \\
SkinToneCCV & & Cheeks-Nose & 0.290 & 0.666 & 0.294 & 0.686 \\
SkinToneNet & & Skin & 0.409 & 0.756 & 0.415& 0.801 \\
SkinToneNet & & Full image & \textbf{0.449} & \textbf{0.901} & \textbf{0.400} & \textbf{0.851} \\\hline
\end{tabular}\vspace{0.3cm}
\caption{\textbf{Effect of split protocol and input representation.} Performance under the STD split is substantially inflated and drops on the identity-held-out Custom Test set, revealing identity leakage. The IND split provides a more reliable estimate of generalization. SkinToneCCV performs best with segmented facial inputs, whereas SkinToneNet performs best with full-image inputs.}
\label{tab:image_ind}
\end{table}

In contrast, the IND split yields lower but more reliable \acrshort{bacc}, with SkinToneNet maintaining strong \acrshort{wooacc} comparable to human off-by-one agreement. This confirms that identity-level separation is essential for evaluating in-the-wild skin tone classification. Consistent with \cite{matias2024enhancing}, which shows that models trained with identity leakage may attend to identity-specific or background cues, all evaluations use the IND split. Regarding input representation, SkinToneCCV performs best with the broader segmented skin region rather than cheeks and nose alone. SkinToneNet performs best with full-image inputs, improving over segmented skin by roughly 5 points in \acrshort{bacc} and 15 points in \acrshort{wooacc}, suggesting that global context helps separate illumination and acquisition effects from intrinsic skin tone.
\begin{table}[h!]
\centering
\small
\setlength{\tabcolsep}{4pt}
\begin{tabular}{l  cc  cc  cc}
\hline
\textbf{Network architecture} & \multicolumn{2}{c}{\textbf{\acrshort{stw}}} & \multicolumn{2}{c}{MSTE} & \multicolumn{2}{c}{CCv2} \\
 & bAcc $\uparrow$ & wOOAcc $\uparrow$ & Acc $\uparrow$ & OOAcc $\uparrow$ & Acc $\uparrow$ & OOAcc $\uparrow$ \\ \hline
LabNet~\cite{robin2020beyond} & 0.328 & 0.777 & 0.227 & 0.588 & 0.195 & 0.507 \\
VehicleNet~\cite{rachmadi2015vehicle} & 0.355 & 0.769 & 0.151 & 0.638 & 0.163 & 0.470 \\
ResNet18~\cite{he2016deep} & 0.389 & 0.856 & 0.321 & 0.740 & 0.228 & 0.615 \\
DenseNet121~\cite{huang2017densely} & 0.445 & 0.860 & 0.292 & 0.792 & 0.227 & 0.626 \\
DINOv2~\cite{oquab2024dinov2learningrobustvisualdinov2} & 0.373 & 0.820 & 0.329 & 0.766 & \textbf{0.276} & 0.650 \\
SkinToneNet (ViT-s) & \textbf{0.449} & \textbf{0.901} & \textbf{0.413} & \textbf{0.853} & 0.250 & \textbf{0.706} \\ \hline
\end{tabular}
\vspace{0.3cm}
\caption{\textbf{Architecture comparison.} SkinToneNet, a fine-tuned ViT-S/16, outperforms baseline architectures and shows strong generalization across STW, MSTE, and CCv2.}
\vspace{-0.5cm}
\label{tab:results_deep_learning}
\end{table}

\paragraph{Architectural comparisons and SkinToneNet.}
As detailed in \autoref{tab:results_deep_learning}, architectures with stronger visual representations, such as DenseNet121, DINOv2, and ViT-S, outperform standard convolutional baselines and prior color-specific networks. SkinToneNet achieves the best \acrshort{wooacc} on \acrshort{stw} and the best \acrshort{ooacc} on MSTE and CCv2, showing that a simple ViT-S/16 fine-tuned on \acrshort{stw} provides a strong reference model for in-the-wild \acrshort{mst} classification. Its full-image input allows the model to use global contextual cues, such as illumination on nearby skin regions, which can support more robust color constancy under uncontrolled acquisition.

\begin{figure}[h!]
    \centering 
    \includegraphics[width=\linewidth]{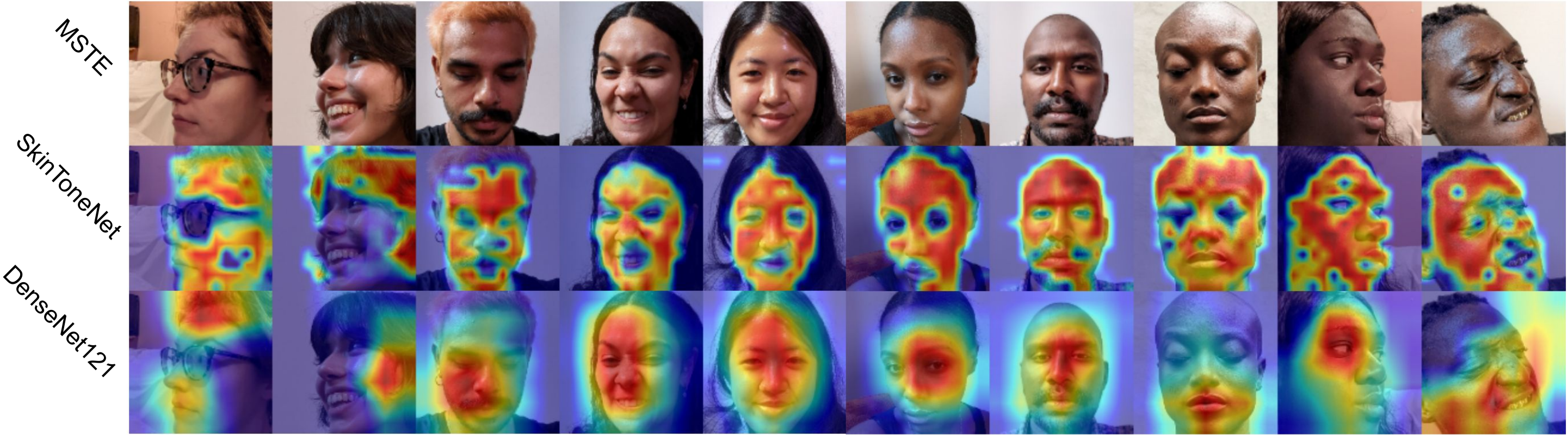}
    \caption{\textbf{Grad-CAM visualizations on the out-of-domain MSTE-G dataset.} The top row shows unconstrained input faces spanning MST tones 1--10. The middle and bottom rows show activation maps for SkinToneNet and DenseNet121, respectively. SkinToneNet produces more localized activations on facial and nearby skin regions.}
    \label{fig:grad_cam}   
\end{figure}

\paragraph{Interpretability and robustness.}
Full-image inputs may introduce shortcut-learning risks, since models could exploit background or identity-specific cues. To diagnose this behavior, we apply GradCAM~\cite{selvaraju2017grad} to the final convolutional and attention layers (\autoref{fig:grad_cam}). SkinToneNet maintains high \acrshort{ooacc} on out-of-domain datasets, and its GradCAM maps concentrate on facial and nearby skin regions across MST 1--10. Compared with the more diffuse activations of DenseNet121, SkinToneNet produces more localized responses, suggesting that its gains are primarily associated with relevant facial cues rather than background artifacts.

\begin{wrapfigure}{r}{0.3\textwidth}
    \vspace{-0.2cm}
\centering
    \includegraphics[width=1\linewidth]{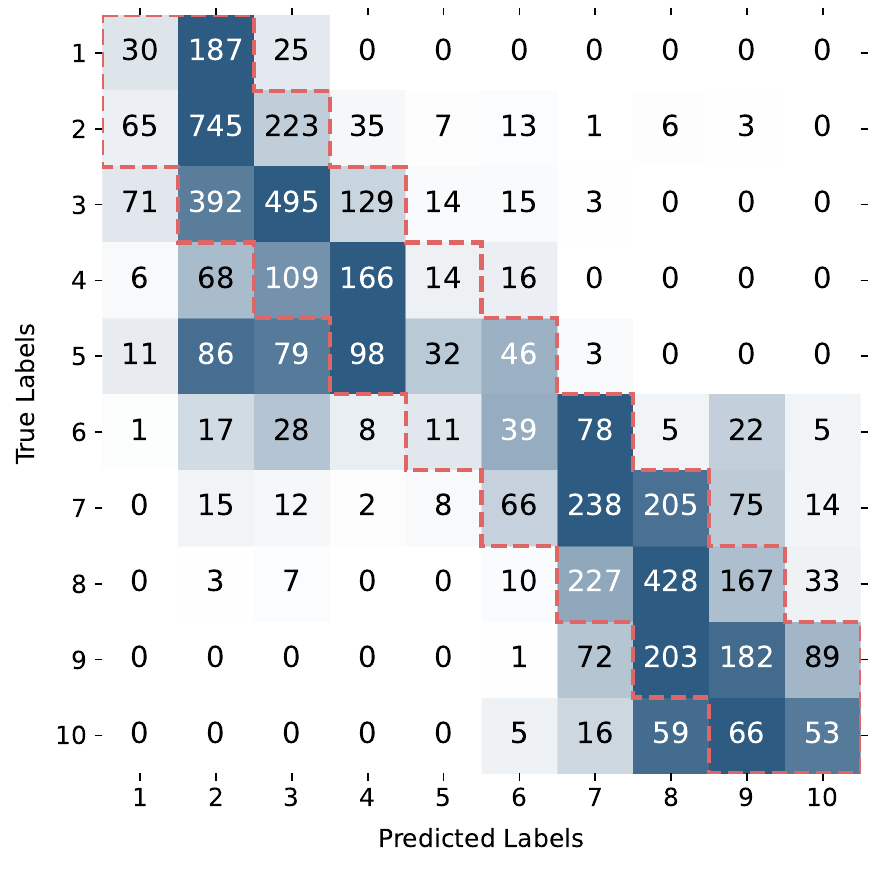}
    \vspace{-0.7cm}
    \caption{Confusion matrix analysis of the SkinToneCCV model. The model collapses towards frequent labels.}
    \label{fig:ccvm_dist}
    \vspace{-0.6cm}
\end{wrapfigure}

\begin{figure}[h]
\centering
    \includegraphics[width=0.75\linewidth]{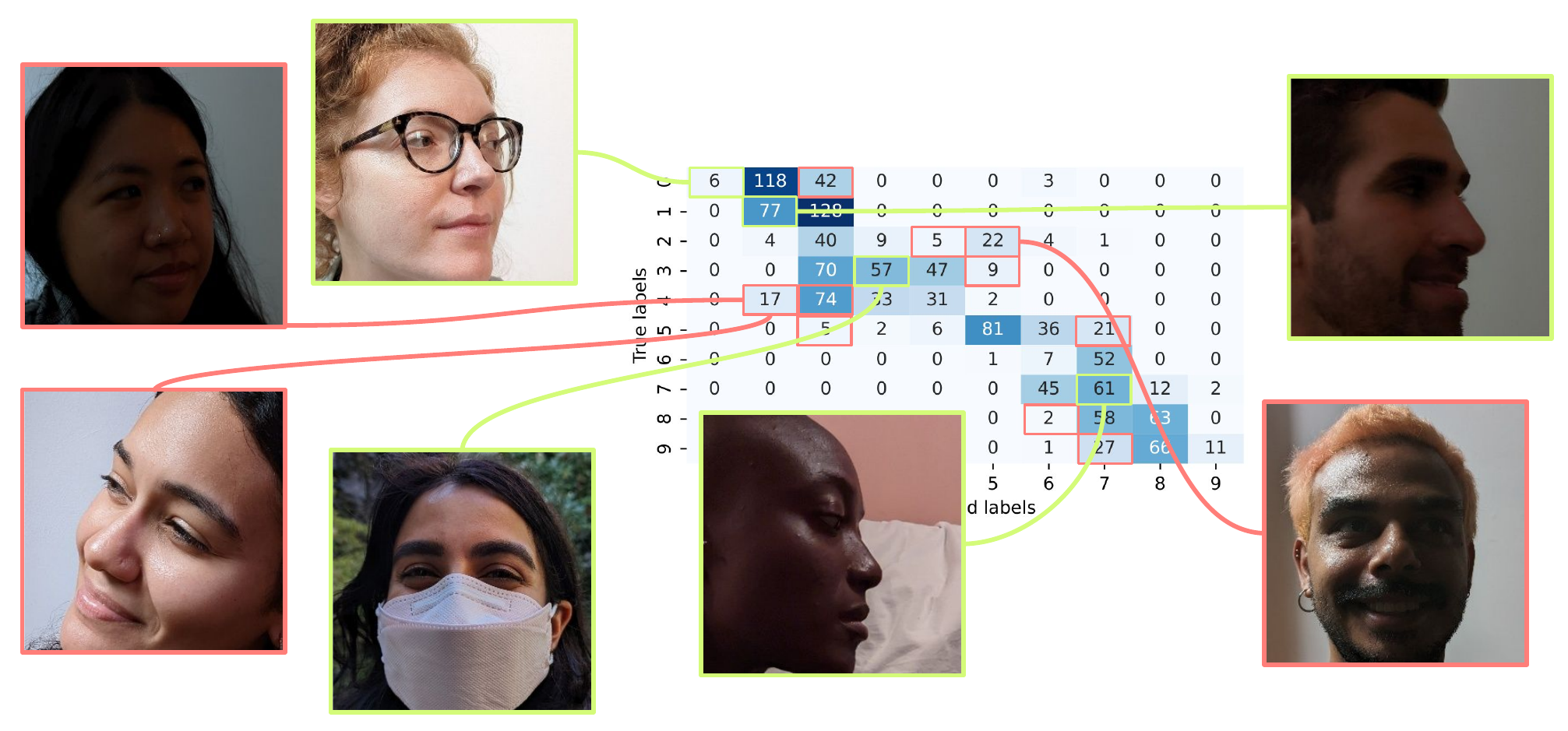}
    \caption{\textbf{Qualitative failure analysis on the MSTE set.} Representative correct predictions are indicated by green borders. Failure cases, defined as instances misclassified by two or more classes, are highlighted with red borders and mapped to their corresponding positions on the confusion matrix.}
    \label{fig:failure}
\end{figure}

\paragraph{Failure analysis.}
As illustrated in \autoref{fig:failure}, SkinToneNet remains robust under several challenging in-the-wild conditions, including face masks and large head-pose variation. Nevertheless, failures occur under complex combinations of occlusion, hair, glasses, tilted head poses, and severe illumination changes. Extreme lighting conditions, such as harsh shadows, overexposure, underexposure, or strong color-temperature shifts, can substantially alter apparent skin reflectance and shift predictions along the \acrshort{mst} scale. While our augmentation strategy mitigates these effects, fully disentangling intrinsic skin tone from extrinsic illumination remains an open challenge.

While SkinToneNet gives the strongest overall results, the behavior of SkinToneCCV is informative for understanding the difficulty of the benchmark. In particular, it isolates how far these carefully engineered hand crafted feature descriptors can go without learning intrinsic feature representations.

\paragraph{SkinToneCCV and baselines.}
Among the evaluated classical variants, SkinToneCCV achieves its best performance using segmented facial inputs, a Random Forest classifier, and a 16-bin rebinarized feature vector. However, as shown in \autoref{fig:ccvm_dist}, the model collapses toward frequent labels, especially MST 2 and 7. This behavior persists even with class-balancing strategies, suggesting that hand-crafted spatial-color statistics are insufficient to disentangle intrinsic skin tone from the illumination, pose, and acquisition variability present in in-the-wild images. Despite this limitation, SkinToneCCV improves substantially over prior classical baselines and reaches performance comparable to earlier learning-based models such as VehicleNet and LabNet on \acrshort{stw}.

\subsection{Auditing widely used facial datasets.}

\begin{figure}[h!]
    \centering
    \includegraphics[width=0.95\linewidth]{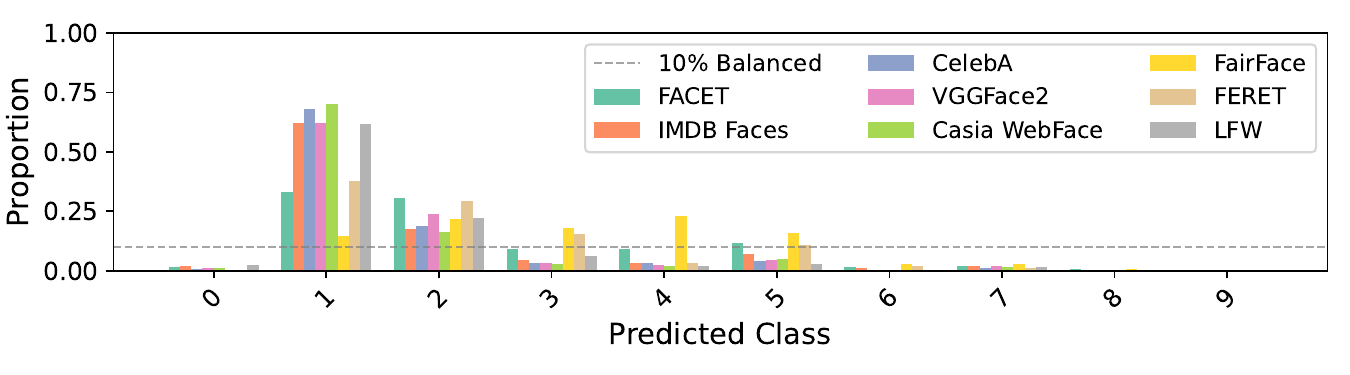}
    \vspace{-0.5cm}
    \caption{\textbf{Zero-shot skin tone distribution audit across standard datasets} \cite{FACET,wang2018devil,liu2018large,cao2018vggface2datasetrecognisingfaces,banerjee2020hallucinating,karkkainen2021fairface,phillips1998feret,huang2008labeled}. Proportions of predicted \acrshort{mst} classes across eight standard vision benchmarks. The dashed horizontal line indicates an ideal uniform distribution of 10\% per class.}
    \label{fig:external_distribution}
\end{figure}

Human facial datasets serve as the foundational bedrock for downstream tasks ranging from face recognition to age regression and facial expression analysis. To quantify demographic representativeness, we deployed \method{} in a zero-shot capacity to audit the presence of the 10 \acrshort{mst} classes across several widely adopted benchmarks, as illustrated in \autoref{fig:external_distribution}.

While datasets specifically curated for fairness evaluation (\eg FACET and FairFace) maintain a reliable proportion of intermediate tones (\acrshort{mst} 3, 4, and 5), the broader landscape exhibits severe demographic imbalances. Crucially, the predicted distributions display a skew toward lighter phenotypes. Furthermore, no evaluated dataset demonstrates a significant presence of darker skin tones (\acrshort{mst} 6 through 10), and the absolute lowest end of the spectrum (\acrshort{mst} 1) is similarly underrepresented. This systemic imbalance means that even benchmarks explicitly built for fairness auditing fail to approximate an ideal, uniform 10\% representation per class. Consequently, these findings expose a critical vulnerability that undermines foundational fairness evaluations across the computer vision community.

\section{Conclusion}

We introduced \acrfull{stw}, a large-scale open-access dataset for \acrlong{itw} skin tone classification on the 10-tone Monk Skin Tone (\acrshort{mst}) scale. \acrshort{stw} contains 42,313 images from 3,564 individuals, improves coverage across the \acrshort{mst} spectrum, and provides reliable annotations supported by strong inter-annotator agreement. Alongside the dataset, we establish a rigorous benchmark protocol based on identity-level splits, ordinal-aware metrics, in-domain evaluation, and cross-dataset generalization.

Our experiments show that skin tone classification remains challenging under unconstrained acquisition conditions. Classical color-descriptor pipelines fail to generalize reliably, while prior learning-based methods perform only modestly on the 10-tone \acrshort{mst} task. In contrast, SkinToneNet, a ViT-S/16 model fine-tuned on \acrshort{stw}, achieves the strongest 10-class performance across \acrshort{stw} and external \acrshort{mst}-annotated benchmarks, including CCv2, MSTE, MSTE-G, and FACET. These results demonstrate the importance of diverse in-the-wild supervision and strict identity-disjoint evaluation for robust skin tone recognition.

Finally, we use SkinToneNet as a zero-shot auditing tool for widely used face datasets, revealing substantial imbalance across \acrshort{mst} classes. In particular, darker tones remain strongly underrepresented, even in datasets designed for fairness evaluation. These findings highlight \acrshort{stw} as both a training resource and a benchmark for measuring skin tone representation in computer vision.

\vspace{0.2cm}
\noindent\textbf{Future work.}
Future directions include extending skin tone auditing to broader media, such as video and educational or textual resources, and further improving the balance of \acrshort{stw} by collecting additional individuals in underrepresented \acrshort{mst} classes.

\bibliography{egbib}
\end{document}